\documentclass[final]{cvpr}

\usepackage{times}
\usepackage{epsfig}
\usepackage{graphicx}
\usepackage{amsmath}
\usepackage{amssymb}

\usepackage{booktabs,amsfonts,dcolumn}
\usepackage{float}
\usepackage{comment}
\usepackage{array}
\usepackage{enumitem}

\usepackage{epstopdf}
\usepackage{multirow}
\usepackage[table]{xcolor}
\usepackage[switch]{lineno}  %
\usepackage{subfigure}

\usepackage{epstopdf}
\usepackage{multirow}
\usepackage{array}
\usepackage[linesnumbered,ruled]{algorithm2e}
\newcolumntype{L}[1]{>{\raggedright\let\newline\\\arraybackslash\hspace{0pt}}m{#1}}
\newcolumntype{C}[1]{>{\centering\let\newline\\\arraybackslash\hspace{0pt}}m{#1}}
\newcolumntype{R}[1]{>{\raggedleft\let\newline\\\arraybackslash\hspace{0pt}}m{#1}}
\usepackage[utf8]{inputenc}
\usepackage[english]{babel}
\usepackage{amsthm}
\usepackage[misc]{ifsym}

\usepackage[table]{xcolor}

\makeatletter\renewcommand\paragraph{\@startsection{paragraph}{4}{\z@}
  {.5em \@plus1ex \@minus.2ex}{-.5em}{\normalfont\normalsize\bfseries}}\makeatother

\newcolumntype{x}[1]{>{\centering\arraybackslash}p{#1pt}}
\newlength\savewidth\newcommand\shline{\noalign{\global\savewidth\arrayrulewidth\global\arrayrulewidth 1pt}\hline\noalign{\global\arrayrulewidth\savewidth}}

\usepackage[pagebackref=true,breaklinks=true,colorlinks,bookmarks=false]{hyperref}

\def\cvprPaperID{416} 
\def\confYear{CVPR 2021}

\begin{document}

\title{Can Temporal Information Help with Contrastive Self-Supervised Learning?}




\author{Yutong Bai\textsuperscript{1$\ast$}~~~~
Haoqi Fan\textsuperscript{2}~~~~
Ishan Misra\textsuperscript{2}~~~~
Ganesh Venkatesh\textsuperscript{3}~~~~
Yongyi Lu\textsuperscript{1 \Letter}~~~~\\
Yuyin Zhou\textsuperscript{1}~~~~
Qihang Yu\textsuperscript{1}~~~~
Vikas Chandra\textsuperscript{3}~~~~
Alan Yuille\textsuperscript{1}\\
\textsuperscript{1} The Johns Hopkins University \qquad
\textsuperscript{2} Facebook AI Research\qquad
\textsuperscript{3} Facebook Reality Labs\\
{\tt\small \{ytongbai, yylu1989, zhouyuyiner, yucornetto, alan.l.yuille\}@gmail.com}\\ {\tt\small \{haoqifan, imisra, gven, vchandra\}@fb.com}
}

\maketitle

\begin{abstract}

Leveraging temporal information has been regarded as essential for developing video understanding models. However, how to properly incorporate temporal information into the recent successful instance discrimination based contrastive self-supervised learning (CSL) framework remains unclear. 
As an intuitive solution, we find that directly applying temporal augmentations does not help, or even impair video CSL in general.
This counter-intuitive observation motivates us to re-design existing video CSL frameworks, for better integration of temporal knowledge.

To this end, we present \textbf{T}emporal-\textbf{a}ware \textbf{Co}ntrastive self-supervised learning (\textbf{TaCo}), as a general paradigm to enhance video CSL. Specifically, TaCo selects a set of temporal transformations not only as strong data augmentation but also to constitute extra self-supervision for video understanding.
By jointly contrasting instances with enriched temporal transformations and learning these transformations as self-supervised signals, TaCo can significantly enhance unsupervised video representation learning.

For instance, TaCo demonstrates consistent improvement in downstream classification tasks over a list of backbones and CSL approaches. Our best model achieves 85.1\% (UCF-101) and 51.6\% (HMDB-51) top-1 accuracy, which is a 3\% and 2.4\% relative improvement over prior arts.

\let\thefootnote\relax\footnote{$^\ast$Work done during Yutong Bai's internship at Facebook.}
\end{abstract}
\vspace{-0.4cm}
\section{Introduction}
\label{sec:introduction}
\begin{figure}\includegraphics[width=1.0\linewidth]{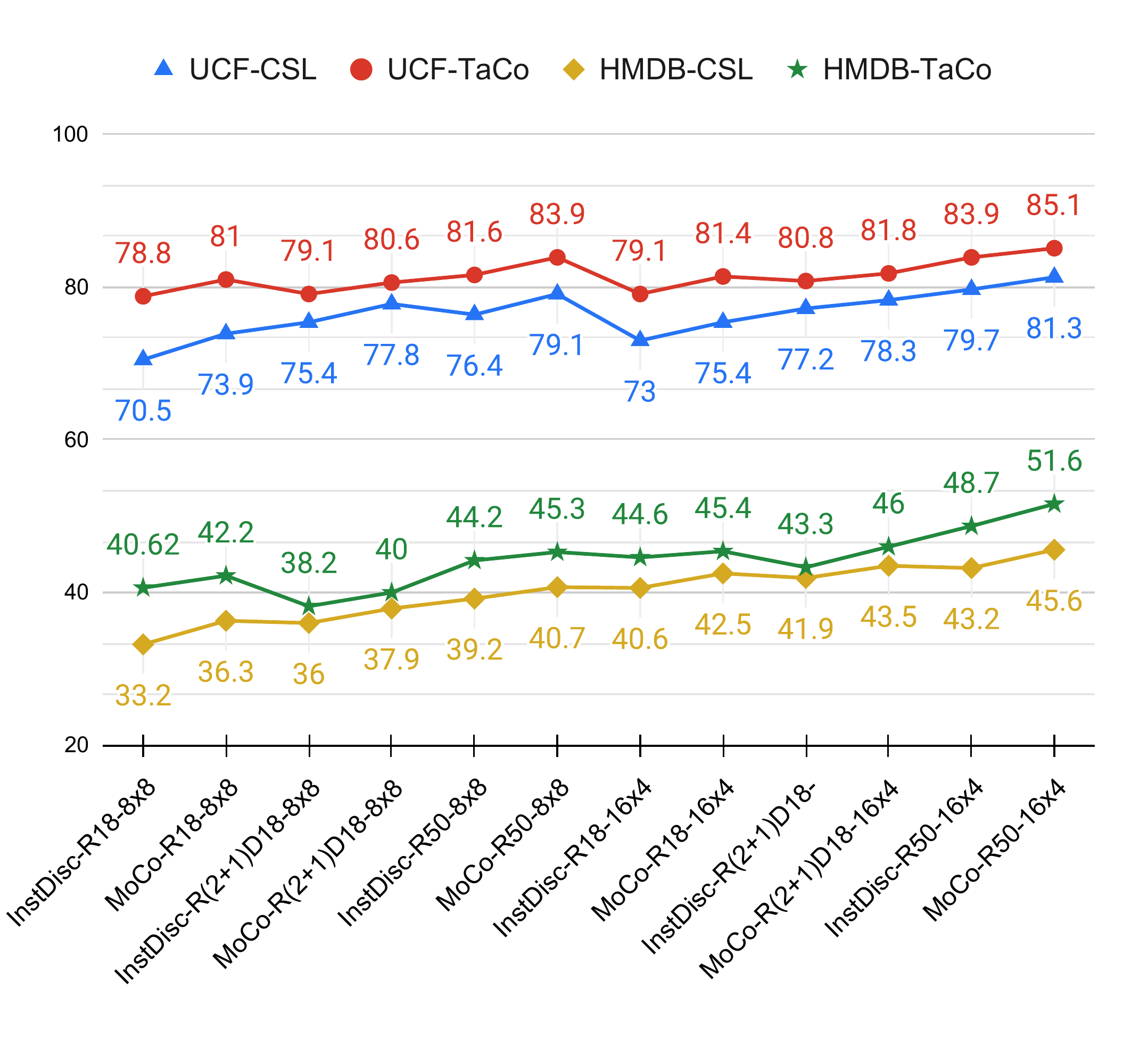}
\caption{\textbf{Top-1 classification accuracy (\%) on UCF-101 and HMDB-51 with CSL and TaCo under different settings.} It contains 2 CSL methods: InstDisc and MoCo; 3 backbone networks: ResNet-18, R(2+1)D-18 and ResNet 50; 2 temporal length: 8 frames with 8 downsampling rate (8$\times$8) and 16 frames with 4 downsampling rate (16$\times$4). TaCo significantly improves the finetuning results compared with vanilla CSL across different settings.}
\label{fig:teaser-image}
\end{figure}

The field of unsupervised representation learning has drawn increasing attention due to the recent success of self-supervised learning techniques. In these approaches, representations are learned through pretext tasks where data itself provides supervision. Typical examples of pretext tasks include recognizing rotation of the image \cite{gidaris2018unsupervised}, predicting relative patch location \cite{doersch2015unsupervised}, solving Jigsaw puzzle \cite{noroozi2016unsupervised} and counting features \cite{noroozi2017representation}. Meanwhile, contrastive self-supervised learning (CSL), a sub-area in self-supervised learning, has gained more momentum since it demonstrates stronger generalization to downstream tasks.
A leading instantiation is a line of instance discrimination based frameworks \cite{wu2018unsupervised, he2020momentum, misra2020self, chen2020simple} wherein an encoded ``query'' and its matching ``key'' are positive pair if they are data augmentations of the same instance and negative otherwise. 

Beyond the image domain, attempts of self-supervised learning have also advanced unsupervised video representation learning.
As the key difference between 2D images and videos lies in the temporal dimension, earlier efforts are mostly dedicated to designing different pretext tasks for video, such as solving video jigsaw puzzles \cite{noroozi2016unsupervised}, recognizing rotations of video clips \cite{gidaris2018unsupervised}, identifying the order of sequences \cite{wei2018learning} and encouraging temporal consistency \cite{CVPR2019_CycleTime, yang2020video}.  
Built upon CSL, a recent study~\cite{qian2020spatiotemporal} proposes to leverage both temporal and spatial information via using temporally consistent spatial augmentations, which largely improves the performance compared with previous approaches.
However, a detailed discussion on how to incorporate temporal information into CSL remains a missing piece in the existing literature.

This prompts one general question: \textbf{``Can temporal information help CSL?''}, which we seek to address throughout the paper. 
We hereby provide the first rigorous study on the effects of temporal information in CSL by taking a step further and answering the following detailed questions:
\begin{itemize}
\item \textit{Question 1: Can we just resort to adding temporal augmentations\footnote{Throughout this paper, \emph{temporal augmentation} refers to different temporal transformations being used as data augmentation.} with existing CSL frameworks?}
\item \textit{Question 2: Is there a more suitable way to model temporal information and learn a better ``solution'' for video CSL?}
\item \textit{Question 3: Is there any innate relation between different video tasks? How can multiple video tasks help with CSL?}
\end{itemize}

To answer the first question, 
we explore different augmentation strategies in video CSL as follows:
1) \textbf{spatial augmentation}, including standard data augmentations such as random flip, shift, and scale as well as those used in image pretext tasks.
2) \textbf{temporally consistent spatial augmentation} \cite{qian2020spatiotemporal} and 3) \textbf{temporal augmentation} (\eg, reversed sequence \cite{wei2018learning}, shuffled clips \cite{fernando2017self} and sped-up sequence \cite{benaim2020speednet}). 
Our empirical results suggest that both standard spatial augmentation and temporally consistent spatial augmentation can consistently benefit video CSL.
However, directly applying temporal augmentation shows limited improvement, or is even detrimental, \ie, lowering the classification accuracy on downstream task (detailed discussions in Section \ref{sec:expresult}).

This surprising result suggests that applying temporal transformations simply as augmentation might not be an appropriate solution for video CSL.
Therefore, we propose to provide additional supervision for those temporal transformations, so that the learned model can be ``aware of'' the temporal relation during training.
Taking chronological order as an example, the model should not only tell that it is positive pair between query (original sequence) and key (augmented reversed sequence), but also learn to reveal whether it's playing forward or backward.

\begin{figure*}[t]
	\centering
        \includegraphics[width=0.80\textwidth]{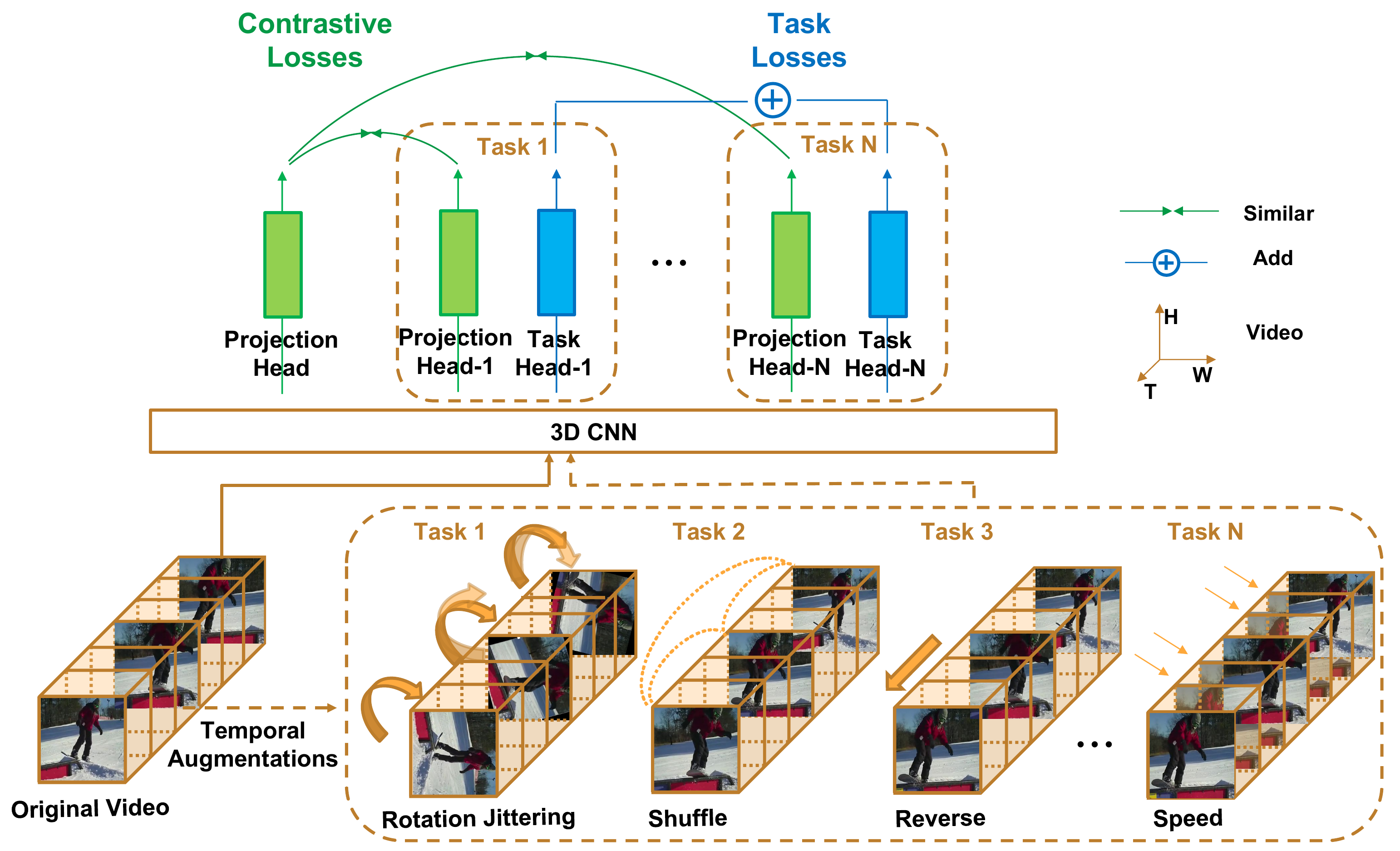}
 \caption{\textbf{Overview of the proposed temporal-aware contrastive self-supervised learning framework (TaCo).} TaCo mainly comprises three modules: temporal augmentation module, contrastive learning module, and temporal pretext task module. For different temporal augmentations, we apply different projection heads and task heads. The features extracted from projection head of original video sequence and augmented sequence are considered as positive sample pairs, and the remaining ones are simply regarded as negative sample pairs. 
 The contrastive loss is computed as the summation of losses over all pairs. }
 \label{fig:framework}
\end{figure*}

To this end, we address the second question and present \textbf{T}emporal-\textbf{a}ware \textbf{Co}ntrastive self-supervised learning (\textbf{TaCo}), as a better solution for video CSL. Specifically, TaCo selects a set of temporal transformations not only as strong data augmentation but also to constitute extra self-supervision under the CSL paradigm. 
What makes TaCo most different from the previous CSL lies in a newly proposed task head besides the projection head for each temporal transformed video. Each task head corresponds to a particular video task and solving those tasks jointly with contrastive learning not only provides stronger surrogate supervision signals during training, but also to learn the shared knowledge among those video tasks (\eg, motion difference between tasks). An overall framework of TaCo is shown in Figure \ref{fig:framework}. 
As a by-product, we provide simple criteria that allows deciding a more suitable task. 

To address the final question, based on the above observation that certain combinations of tasks do perform better, such as \textbf{{\em  shuffle}}  with \textbf{{\em  speed}} task, \textbf{{\em  rotation jittering}}  with \textbf{{\em  reverse}}  task (defined in Section~\ref{sec:temporalpretexttasks}), we are also interested in exploring the multi-task relation. We argue that when tasks harmonize with each other, TaCo with dual-task setting can further boost the performance. Two complementary cues are discussed, \ie, task similarity and task motion difference for measuring the harmony of tasks. Lastly, extensive ablation experiments demonstrate that TaCo also shows its transferring ability among diverse tasks, backbones, and contrastive learning approaches (see Figure~\ref{fig:teaser-image}).

Therein, we conclude that temporal information can indeed help CSL, and identify that the key lies in the extra guidance from temporal augmentations as self-supervised signals.
To summarize,
our contributions are four-fold:
\begin{itemize}
    \item A general video CSL framework called TaCo, which enables effective integration of temporal information.
    \item As a generic and flexible framework, TaCo can well accommodate various temporal transformations, backbones, and CSL approaches.
    \item Based on TaCo, we provide the first rigorous study on the effects of temporal information in video CSL.
    \item TaCo achieves  85.1\% (UCF-101) and 81.3\% (HMDB-51) top-1 accuracy, which is a 3\% and 2.4\% relative improvement over prior arts.
\end{itemize}
\section{Related Work}
\label{sec:relatedwork}
\paragraph{Contrastive Self-Supervised Learning.}CSL has demonstrated its great potential in learning visual representation from unlabeled data. Instance Discrimination (InstDisc) \cite{wu2018unsupervised} provides an early CSL framework to learn feature representation that discriminates among individual instances using a memory bank. It has been extended in many recent works \cite{he2020momentum, misra2020self, tian2019contrastive, chen2020simple}. MoCo \cite{he2020momentum} structures the dictionary as a queue of encoded representations. PIRL \cite{misra2020self} focuses on learning invariant representation of the augmented view. SimCLR \cite{chen2020simple} studies CSL components and shows the effects
of different design choices. We show that the proposed TaCo can well accommodate different CSL approaches.

\paragraph{Pretext Tasks for Image.} Early works on self-supervised learning mainly focus on designing handcrafted pretext tasks. Examples such as predicting the rotation angle of the transformed images \cite{gidaris2018unsupervised}, predicting the relative position \cite{doersch2015unsupervised}, Jigsaw puzzle \cite{noroozi2016unsupervised}, counting features \cite{noroozi2017representation} and colorization \cite{zhang2016colorful}. Instead of spatial transformation, we explore temporal transformation in this paper. 

\paragraph{Pretext Tasks for Video.} Among all the video pretext tasks, some are directly extended from image, such as predicting the rotation video clip \cite{jing2018self}, solving cubic puzzles \cite{kim2019self}. Compared to image counterpart, video can provide extra temporal supervision signals. The learning of video representation can be driven by validating sequence ordering of the video clips \cite{xu2019self}, by detecting the shuffled clips \cite{fernando2017self} and by predicting if a video clips is at its normal frame rate, or down-sampled rate \cite{benaim2020speednet}. Recent works use temporal correspondence as a supervision signal \cite{CVPR2019_CycleTime, yang2020video, purushwalkam2020aligning}. 

\paragraph{Multi-Task Learning} Our work is also related to multi-task learning. Recent work \cite{doersch2017multi} combines four pretext tasks for self-supervised learning. \cite{taleb20203d} extends the multiple pretext tasks learning to medical imaging. However, none of these architectures are designated for video.

\section{Methodology: Temporal-aware Contrastive Self-supervised Learning (TaCo)}
\label{sec:method}

\begin{figure}[t]
\centering
\includegraphics[width=0.6\linewidth]{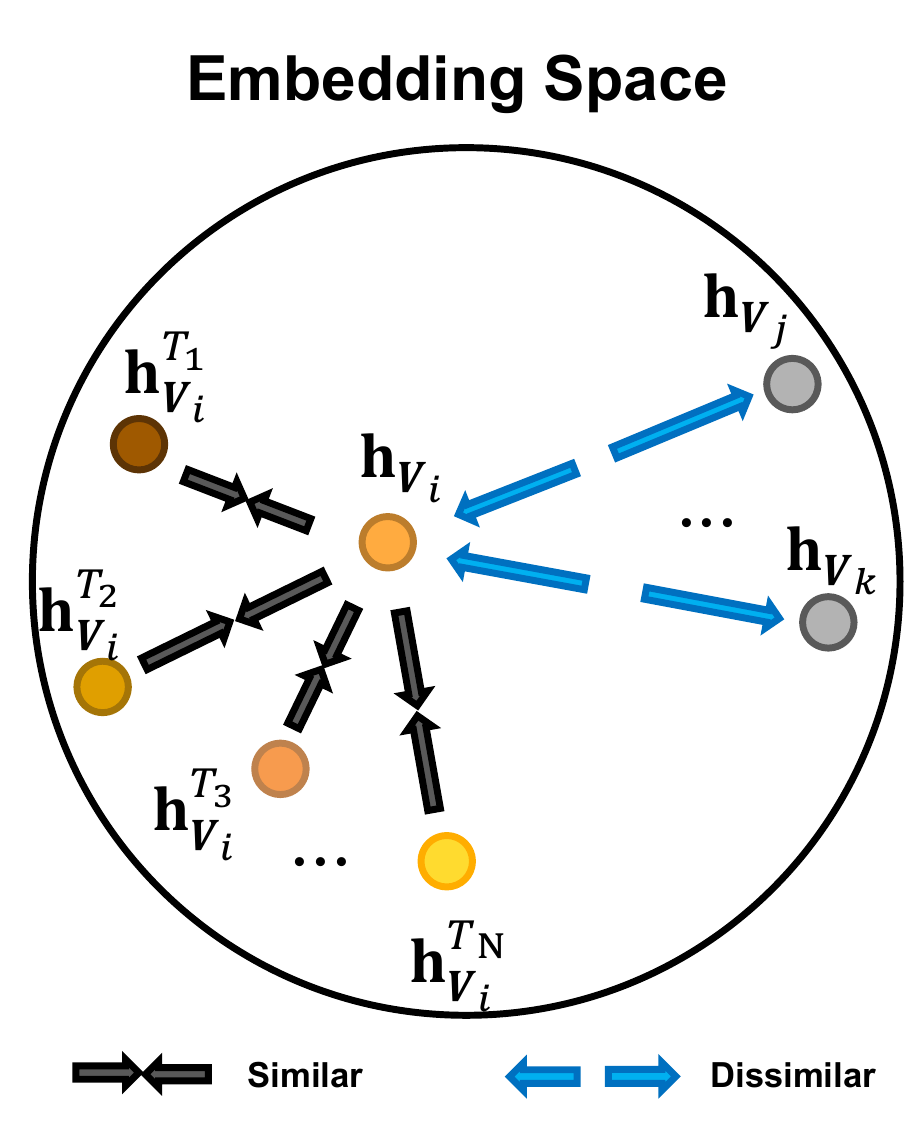}
\caption{\textbf{The illustration for contrastive losses}. For video $\mathbf{V}_i$ and temporal augmented videos $\{\mathbf{V}^{T_n}_i\}$, we use encoder and projection head to extract the feature representation $f_i$ and $\mathbf{h_V}^{T_n}_i$, $n \in [1, N]$ (N is the number of task). Colored dots represent  videos' feature after projection head. It encourages original feature $\mathbf{h_V}_i$ to be similar with those of its temporal augmented video features $\mathbf{h_V}^{T_n}_i$, $n \in [1, N]$. Gray dots represent features extracted from other videos, which we encourage original feature $\mathbf{h_V}_i$ to be dissimilar with.
}

\label{fig:embedding} 
\end{figure}

As illustrated in Figure \ref{fig:framework}, the proposed Temporal-aware Contrastive self-supervised learning framework (TaCo), is comprised of the following three major modules: temporal augmentation module, contrastive learning module, and temporal pretext task module. 
Given a video sequence, we first apply temporal transformations to obtain a set of augmented video sequences. 
For the original video, we use a projection head for feature extraction. 
For the augmented video sequences, we apply two heads, one projection head and one task head. Projection head is used to extract feature for contrastive learning and task head is used for solving the pretext tasks to provide extra self-supervised signals. The features extracted from the projection head of the original video as well as the augmented video sequence are considered as positive pairs while the remaining pairs are regarded as negative samples (Figure~\ref{fig:embedding}). 
The overall objective is the weighted sum of the contrastive loss and the task loss.

\subsection{Temporal Augmentation Module}
\begin{itemize}
    \item A strong \textit{temporal augmentation} module. Suppose we are given a set of videos, $\mathcal{D} = \{\mathbf{V}_1, \dots, \mathbf{V}_{K}\}$, $K = | \mathcal{D} |$,  with $\mathbf{V}_i \in \mathbb{R}^{T \times H \times W \times 3}$, and a set of strong temporal augmentations, $\mathcal{T} = \{T_1, T_2, \dots, T_N\}$, where $N = |\mathcal{T}|$. Given an video $\mathbf{V}_i$ , we first apply temporal augmentations on the video: $\{ \mathbf{V}^{T_1}_i, \mathbf{V}^{T_2}_i, ...\mathbf{V}^{T_N}_i \}$. Here we analyze temporal augmentations, including temporal \textbf{{\em  rotation jittering}}, \textbf{{\em  shuffle}}, \textbf{{\em  speed}} and \textbf{{\em  reverse}}. Details about these pretext tasks can be found in Section~\ref{sec:temporalpretexttasks}. Other spatial augmentations are discarded without further consideration (not our main focus).
\end{itemize}
\subsection{Contrastive Learning Module}
\begin{itemize}
    \item A neural network \textit{base encoder} $f(\cdot)$ that extracts representation vectors for both original video and temporal augmented videos. 
    Our framework allows various choices of the network architecture without any constraints.
    $\mathbf{h_{V}} = f(\mathbf{V})$, where $\mathbf{h_{V}}$ is the output after the average pooling layer. 
    
    \item A \textit{projection head} $g(\cdot)$ that maps representations to the space where contrastive loss is applied. 
    We use a linear layer to obtain $\mathbf{z_V} = g(\mathbf{h_{V}})=W^{(1)}\mathbf{h_{V}}$, followed by a $\ell_2$ normalization layer. MLP is also applicable here. In this case, $\mathbf{z_V} = g( \mathbf{h_{V}}) =W^{(2)}\sigma(W^{(1)}\mathbf{h_{V}})$ where $\sigma$ is a ReLU non-linearity. It is worth noting that our $\mathbf{z_V}$ here is different from previous contrastive learning's embedding feature since we use extra guidance provided by task loss, which would be elaborated on in Section \ref{sec:taskhead}.
    \item \textit{Contrastive loss}. Let $\mathrm{sim}(\boldsymbol u,\boldsymbol v) = \boldsymbol u^\top \boldsymbol v / \lVert\boldsymbol u\rVert \lVert\boldsymbol v\rVert$, which denotes the dot product between $\ell_2$ normalized $\boldsymbol u$ and $\boldsymbol v$ ($\ie$ cosine similarity). We consider the temporal augmented video $\mathbf{V}^{T_n}$ as positive samples to $\mathbf{V}$. The loss function for a single positive pair of examples $(\mathbf{V}_i, \mathbf{V}_i^{T_n})$ is defined as a NCE loss \cite{cpc, gutmann2010noise}
    \begin{equation}
    \label{eq:loss}
        \ell_{\mathbf{V}_i, \mathbf{V}_i^{T_n}} = -\log \frac{\exp(\mathrm{sim}(\mathbf{z}_{\mathbf{V}_i^{T_n}},  \mathbf{z}_{{\mathbf{V}_i}})/\tau)}{\sum_{j=0}^{K} \exp(\mathrm{sim}(\mathbf{z}_{\mathbf{V}_i^{T_n}},  \mathbf{z}_{\mathbf{V}_j})/\tau)}~,
    \end{equation}
    where $\tau$ denotes a temperature parameter. This loss encourages the representation of video $\mathbf{V}$ to be similar to that of its temporal augmented video counterpart $\mathbf{V^{T_n}}$, while dissimilar to the representation of other videos.
    The overall contrastive loss is computed across all positive pairs:
    \begin{equation}
        \ell_{contrast}(\mathbf{V}) = \ell_{\mathbf{V}, \mathbf{V}^{T_1}}+\ell_{\mathbf{V}, \mathbf{V}^{T_2}}+ ... +\ell_{\mathbf{V}, \mathbf{V}^{T_N}}.
    \end{equation}
\end{itemize}

\subsection{Temporal Pretext Task Module}
\label{sec:taskhead}
\begin{itemize}
    \item A \textit{task head} $t(\cdot)$ for solving pretext tasks. For different tasks, we apply different task heads,  The details of these design are discussed more in Section~\ref{sec:temporalpretexttasks}. 
    \item A \textit{task loss} as a self-supervised signal, defined for a pretext task. Task loss works as one of the self-supervised signals. $t(\mathbf{h}^{T_n}_{\mathbf{V}}))$ followed by a softmax layer is used to solve the pretext task as a classification problem. For different temporal related tasks, we simply set all weights to be equal as in Equation~\ref{eqn:task-related-losses}, since they are generally considered as equally important for incorporating temporal information.
    \begin{equation}
        L_{task} = L_{task_1} + L_{task_2} + ... + L_{task_N}
    \label{eqn:task-related-losses}
    \end{equation}
\end{itemize}

\subsection{Overall Loss Function.}
The overall loss function can be formulated as:
\begin{equation}
    L_{overall} = L_{contrast} + \lambda L_{task},
\label{eqn:overall-loss}
\end{equation}
where $\lambda$ is used for balancing the two loss terms. 
Note that a larger $\lambda$ is generally required here since the numerical value of 
$L_{contrast}$, \ie the loss summation over excessive pairs, is usually much larger than that of $L_{task}$.

\section{Experiments}
\label{sec:experiments}
\subsection{Datasets}
The whole experiments contain three parts: self-supervised pretraining, supervised fine-tuning and testing. 
We conduct TaCo pretraining on Kinetics 400 dataset \cite{kay2017kinetics}, which contains 306k 10-second
video clips covering 400 human actions. During unsupervised pretraining, labels are dropped. For fine-tuning and testing, we evaluate on both UCF-101 \cite{soomro2012ucf101} and HMDB-51 \cite{kuehne2011hmdb} datasets, which has 13k videos spanning over 101 human actions and 7k videos spanning over 51 human actions respectively. We extract video frames at the raw frame per second.
\subsection{Self-supervised Pretraining}
\label{sec:exp:self-supervised-pretraining}
We apply 3D-ResNets as backbone to encode a video sequence. We expand the original 2D convolution kernels to 3D to capture spatiotemporal information in videos. The design of our 3D-ResNets follows the “slow” pathway of the SlowFast \cite{feichtenhofer2019slowfast} network with details revealed in PySlowFast codebase \cite{fan2020pyslowfast}. The initial learning rate is set to 0.06. We use SGD as our optimizer with the momentum of 0.9. All models are trained with the mini-batch size of 1024. We linearly warm-up the learning rate in the first 5 epochs followed by the scheduling strategy of half period cosine learning rate decay. In all experiments, 8 consecutive frames are used as the input unless otherwise specified. More specifically, 8-frame clips are sampled with the temporal stride of 8 from each video for the self-supervised pre-training. 
\subsubsection{Contrastive Learning Framework}
\paragraph{InstDisc:} Following~\cite{wu2018unsupervised}, the temperature $\tau$ is set as 0.07 in the InfoNCE loss for all experiments. The number of sampled representation N is 16384.
\paragraph{MoCo:} Following~\cite{he2020momentum}, we use synchronized batch norm to avoid information leakage or overfitting. Temperature $\tau$ is set to be 0.07 in the InfoNCE loss for all experiments.
\vspace{-0.1cm}
\subsubsection{Temporal Pretext Tasks}
\label{sec:temporalpretexttasks}
\paragraph{Rotation Jittering Task.} Different from the spatial rotation task~\cite{gidaris2018unsupervised} which is originally designed for 2D natural images, here we propose to apply jittered rotation to different temporal frames, termed as \textbf{{\em  Rotation Jittering}}. 
We note that this simple modification allows us to enjoy additional benefits brought by exploiting information along the temporal dimension.

In our implementation, similar to~\cite{gidaris2018unsupervised}, we define 4 \textbf{{\em  rotation jittering}}  recognition tasks, \ie, the rotation transformations of ~$0^{\circ}, 90^{\circ}, 180^{\circ}, 270^{\circ}$, respectively.
In addition, to ensure that the rotation degrees of all frames differ from each other, we add a random noise degree $\alpha \in [-3^{\circ}, 3^{\circ}]$ to the original rotation angle.
After obtaining the embedded features from the 3D encoder, we design a projection head and a \textbf{{\em  rotation jittering}}  task head for this task specially. 
For the projection head design, we first adopt a fully connected layer, followed by a reshaping operation.
Then the reshaped feature representation is $\ell2$ normalized and fed forward to the final linear layer.
For \textbf{{\em  rotation jittering}}  task head, we first use a fully connected layer. And after reshaping, we adopt one more linear layer with $\ell2$ norm in the end.

\paragraph{Temporal Reverse Task.} Similar to~\cite{wei2018learning}, we formulate temporal direction as self-supervised signals in our temporal \textbf{{\em  reverse}}  task design. Specifically, we aim to predict whether
a video sequence is playing forwards or backwards.
The projection head and \textbf{{\em  reverse}} task head follow similar designs stated above for the \textbf{{\em  rotation jittering}} task. 

\paragraph{Temporal Shuffle Task.} Following \cite{fernando2017self}, for a video with input length $t$, we randomly sample 3 $t/2$-frame clips out of it and choose one to shuffle its frames. Then we predict which clip is shuffled. The projection head design is the same as the projection head design of the \textbf{{\em  rotation jittering}}  task. For \textbf{{\em shuffle}} task head design, following~\cite{fernando2017self}, we apply a fully connected layer for each clip's embedded feature. These layers
from each branch are summed after taking the pair-wise activation difference leading to a $d$ dimensional vector, where $d$ is the dimensionality of the first fully connected layer. We then feed this fused activation vector through two fully connected layers with ReLU between them, followed by a softmax classifier in the end. 

\paragraph{Temporal Speed Task.} Similar to~\cite{benaim2020speednet}, given a video, we apply different sampling rates (\eg, 1$\times$, 2$\times$, 4$\times$, ...) to obtain clips with different playing speed, which we aim to predict in the self-supervised pretraining.
For projection head and \textbf{{\em speed}} task head design, we adopt a similar one as temporal \textbf{{\em shuffle}}  task, to better learn the relatively subtle motion difference.
\subsubsection{Supervised Classification}
Following \cite{wang2019self, jing2018self, kim2019self, benaim2020speednet, xu2019self}, 
the self-supervised pretrained model is then fully finetuned, \ie, finetuning on all layers, for different downstream classification tasks.
In order to conduct a fair comparison, we finetune the classifier on UCF-101 \cite{soomro2012ucf101} and HMDB-51\cite{kuehne2011hmdb} datasets for action recognition. 
The performance is quantified via the standard evaluation protocol in~\cite{wang2018non, feichtenhofer2019slowfast,fan2020pyslowfast}, \ie, uniformly sampling 10 clips of the whole video and averaging the softmax probabilities of all clips as the final prediction. 
We train TaCo for 200 epochs with a batch size of 128 and an initial learning rate of 0.2, which is reduced by a factor of 10 at 50, 100, 150 epoch respectively.
\subsubsection{Linear Evaluation}
We also evaluate the video representations by only finetuning the last linear layer, \ie, the weights of all other layers are fixed during finetuning. 
For linear evaluation, we set the initial learning rate to 30, the weight decay to 0, and the number of total epochs to 60. 
The learning rate is decaying at epoch 30, 40, 50 with the decay rate as 0.1.

\subsection{Discussion}
\label{sec:expresult}
Without loss of generality, we use the 3D InstDisc pipeline \cite{wu2018unsupervised} to illustrate the effectiveness of TaCo.
Our backbone architecture is 3D-ResNet18~\cite{feichtenhofer2019slowfast}.
\paragraph{Can Temporal Augmentations help video CSL?}
We first conduct experiments using temporal augmentations as in PIRL~\cite{misra2020self}. Table \ref{tab:singleTask} shows, surprisingly, that directly applying temporal augmentation brings no obvious improvement over 3D InstDisc. Instead, sometimes it even causes performance drop. 
For example, considering the \emph{fully finetune} setting, we can find slight dip in performance with all the four temporal augmentations compared to the baseline. 
Among them, \textbf{{\em  speed}}  augmentation achieves the best result on UCF-101, with 69.9\% top-1 accuracy, which still lower than 3D InstDisc (70.33\%). Results on HMDB-51 suggest a similar trend. 

\begin{table}[h]
\footnotesize
\centering
\begin{tabular}{c|c|c|c}
\shline
\multicolumn{2}{c|}{\multirow{2}{*}{Method}} &
  \multicolumn{2}{c}{Fully Finetune} \\ \cline{3-4} 
\multicolumn{2}{c|}{}              & UCF   & HMDB\\ \shline
\multicolumn{2}{c|}{baseline (3D InstDisc)} &  70.33          & 33.23 \\ \shline
\multirow{4}{*}{\begin{tabular}[c]{@{}c@{}} Temporal\\ Augment \end{tabular}} &
  Rotation &
  68.61 &
  31.73 \\ \cline{2-4} 
              & Reverse             & 69.27          & 33.01 \\ \cline{2-4} 
              & Shuffle               & 69.41          & 31.87 \\ \cline{2-4} 
              & Speed               & 69.90          & 32.98 \\ \shline
\multirow{4}{*}{\begin{tabular}[c]{@{}c@{}}TaCo\\ (Ours)\end{tabular}} &
  Rotation &
  73.81 &
  36.89\\ \cline{2-4} 
              & Reverse             & 74.57 & \textbf{37.23} \\ \cline{2-4} 
              & Shuffle              & \textbf{76.02} & 35.59 \\ \cline{2-4} 
              & Speed                & 74.38          & 36.13 \\ \hline
\end{tabular}
\caption{Comparing individual video pretext tasks: top-1 accuracy (\%) on UCF-101 and HMDB-51 classification of TaCo using different pretext tasks with fully finetuning schedule. Here we provide the CSL alone results; CSL with temporal augmentation results and TaCo with task loss results.}
\label{tab:singleTask}
\end{table}
\paragraph{Temporal pretext task as self-supervision?}
As can be seen in Table~\ref{tab:singleTask}, by forming the temporal transformation as additional self-supervision, 
our TaCo with different task choices consistently shows significant improvements (by more than 3\% in terms of accuracy for most cases) compared to the baseline. 
This demonstrates that the proposed task loss provides extra self-supervision which enables the integration of temporal information, therefore guides TaCo to learn better video representation.

\begin{figure}[h]
    \centering
    \includegraphics[width=0.75\linewidth]{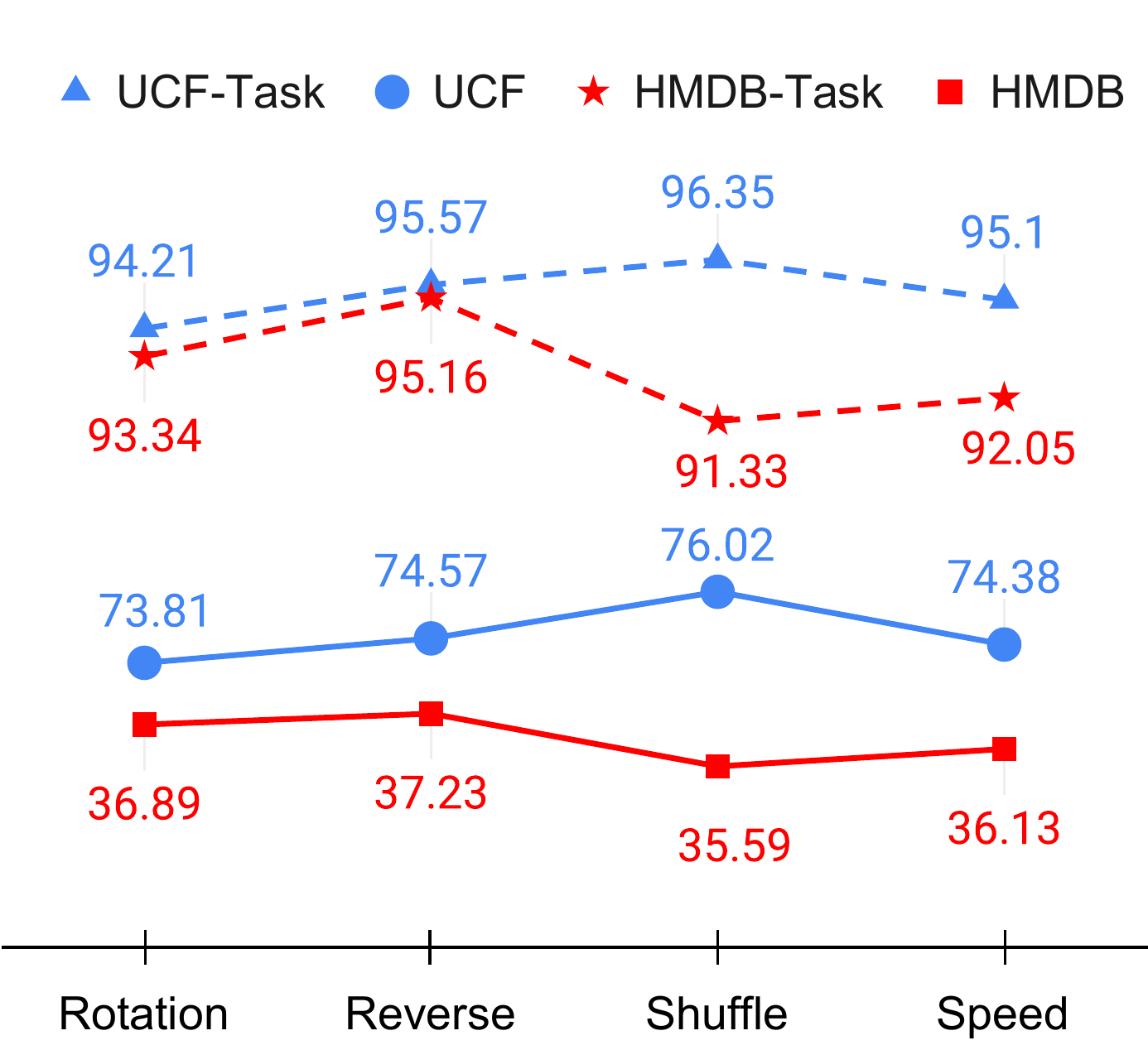}
    \caption{TaCo with different pretext tasks (solid lines) vs. CSL models finetuned on different pretext tasks (dash lines).
    }
    \label{fig:trend}
\end{figure}

\begin{itemize}[leftmargin=*]
   \setlength\itemsep{0.1em}
   \item\textbf{Not all pretext tasks are equally suited to video CSL. } We've shown in Table \ref{tab:singleTask} that each of the four video pretext tasks boosts the contrastive self-supervised learning performance. 
   However, we've found that not all video pretext tasks are suitable for video CSL. 
   For instance, 
   Under the 8 frame setting, our empirical results show a clear performance drop of TaCo by using ClipOrder \cite{xu2019self}.
   We hypothesis that this is due to that 8 frames are too short for extracting effective temporal information from re-ordering the video sequence. However, re-ordering longer clips may result in significantly heavier computation overhead, which can be impractical.
   Therefore, we only select pretext tasks which can extract effective temporal information with practical computation budgets, such as \textbf{{\em  speed}} task or \textbf{{\em  rotation jittering}}  task.
    \item \textbf{A simple solution to select the best pretext task for video CSL.}
    To analyze the correlation between the pretext task and contrastive learning, we perform contrastive learning as pretraining and then finetune all layers on different pretext tasks.
    Intuitively, a higher finetuning performance should indicate a better pretext task to be employed.
    Figure \ref{fig:trend} shows the fine-tuning top-1 accuracy of different video pretext tasks on UCF-101 and HMDB-51, pretrained with 3D InstDisc (in dash lines).
    As a comparison, the results of TaCo on different pretext tasks are also demonstrated (in solid lines). 
    Based on the results,
    we can see that if a task shows better performance during fine-tuning, it also coordinates better with contrastive learning. For example, since \textbf{{\em  shuffle}}  task shows the best fine-tuning accuracy on UCF-101 (96.35\%), we can infer that it will also achieve the best result with TaCo.
    This conjecture coincides with our experimental result (\textbf{{\em  shuffle}}  task achieves the highest accuracy of 76.02\% among all pretext tasks with TaCo).
    Therefore, by using the finetuning performance as an indicator, we can easily select better pretext tasks to be used in TaCo, which in turn largely alleviates the training budget.
\end{itemize}

\begin{figure}[h]
\centering
\includegraphics[width=0.65\linewidth]{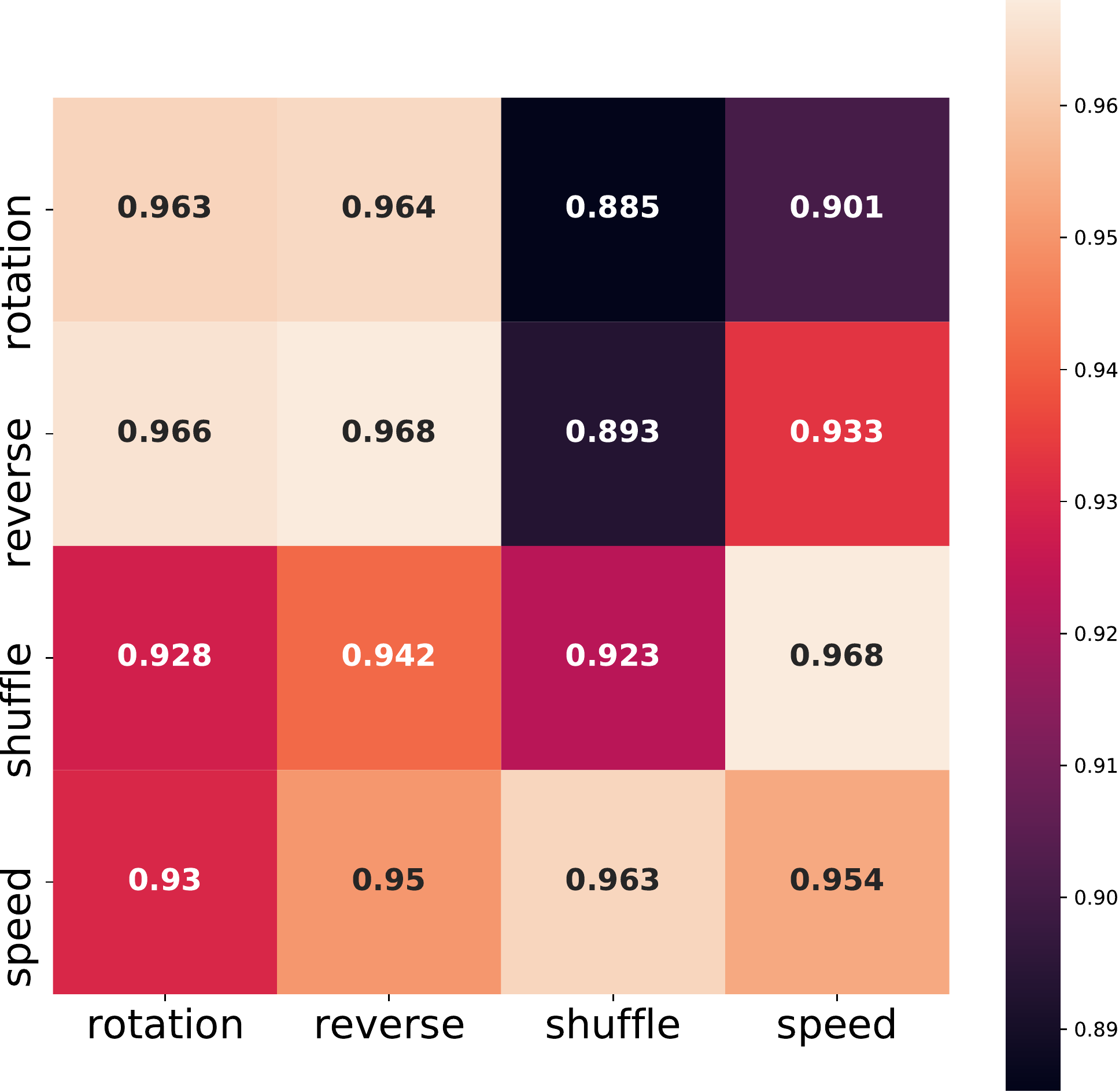}
\caption{The illustration of pretext tasks' relation. The number on each grid indicates the top-1 pretext task prediction accuracy (\%) on UCF-101. Models are pretrained on the tasks each row, and finetuned on each column.}
\label{fig:pretext-task} 
\end{figure}

\paragraph{Multiple pretext tasks?}
\begin{figure*}[t]
    \centering
    \includegraphics[width=1.0\linewidth]{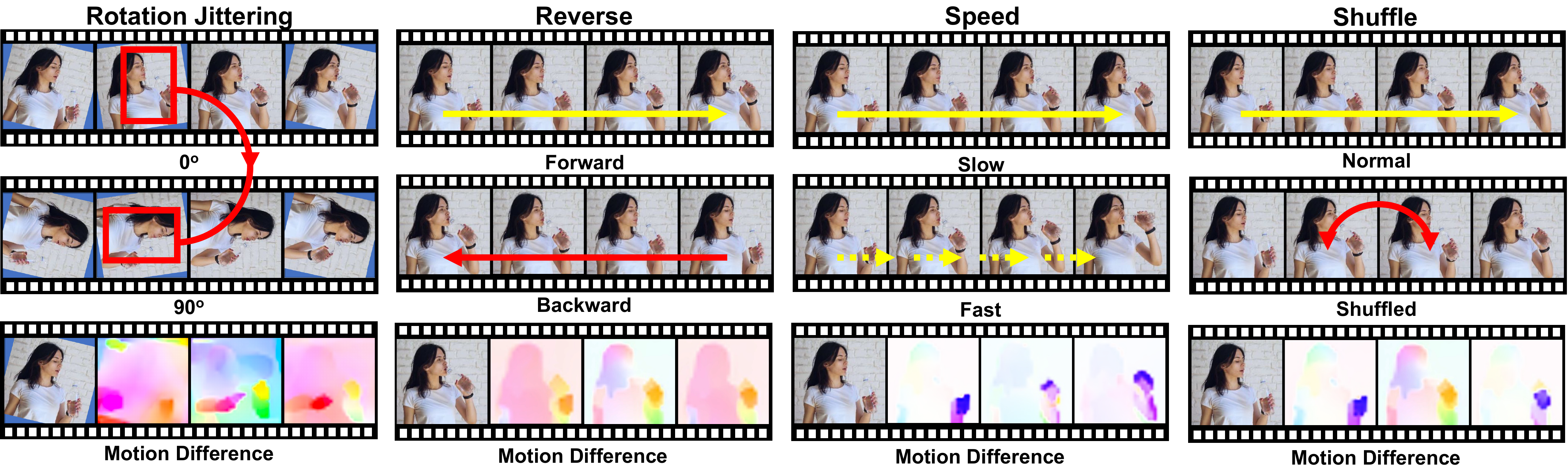}
    \caption{\textbf{Illustration of different temporal pretext tasks with two possible cases and their motion difference.} Left to right: four different kinds of temporal pretext tasks. Top two rows show two possible cases of each task, while last row show the flow field difference between these two cases. Arrows indicate the tasks objective. Best viewed in color and zoomed in. (This video was used for illustrative purposes only.)}
    \label{fig:teaser}
\end{figure*}
\begin{table}[]
\centering
\footnotesize
\begin{tabular}{c|c|c|c|c}
\shline
\multicolumn{3}{c|}{Method}        & UCF            &HMDB\\ \shline
\multicolumn{3}{c|}{CSL: 3D InstDisc} & 70.33          & 33.23          \\ \shline
\multirow{4}{*}{\begin{tabular}[c]{@{}c@{}}Single-\\ Task\end{tabular}} &
  \multirow{4}{*}{\begin{tabular}[c]{@{}c@{}}TaCo\\(Ours)\end{tabular}} &
  Rotation &
  73.81 &
  36.88 \\ \cline{3-5} 
       &      & Reverse             & 74.57          & \textbf{37.23} \\ \cline{3-5} 
       &      & Shuffle              & \textbf{76.02} & 35.59          \\ \cline{3-5} 
       &      & Speed               & 74.38          & 36.13          \\ \shline
\multirow{12}{*}{\begin{tabular}[c]{@{}c@{}}Dual-\\ Task\end{tabular}} &
  \multirow{6}{*}{\begin{tabular}[c]{@{}c@{}} \\Temporal\\Augment  \end{tabular}} &
  Rotation+Reverse &
  70.28 &
  32.04 \\ \cline{3-5} 
       &      & Rotation+Shuffle     & 70.11          & 32.74          \\ \cline{3-5} 
       &      & Rotation+Speed      & 70.19          & 33.01          \\ \cline{3-5} 
       &      & Shuffle+Reverse      & 70.17          & 32.62          \\ \cline{3-5} 
       &      & Speed+Shuffle        & 70.22          & 32.87          \\ \cline{3-5} 
       &      & Reverse+Speed       & 69.99          & 32.10          \\ \cline{2-5} 
 &
  \multirow{6}{*}{\begin{tabular}[c]{@{}c@{}}TaCo\\(Ours)\end{tabular}} &
  Rotation+Reverse &
  75.24 &
  39.02 \\ \cline{3-5} 
       &      & Rotation+Shuffle     & 74.92          & 37.44          \\ \cline{3-5} 
       &      & Rotation+Speed      & 75.03          & 38.15          \\ \cline{3-5} 
       &      & Shuffle+Reverse      & 75.69          & 36.54          \\ \cline{3-5} 
       &      & Speed+Shuffle        & \textbf{78.77} & \textbf{40.62} \\ \cline{3-5} 
       &      & Reverse+Speed       & 75.18          & 38.96          \\ \hline
\end{tabular}
\caption{{Comparing combination of video pretext tasks:} top-1 accuracy (\%) of the classification results on UCF-101 and HMDB-51 with fully finetuned backbone of TaCo.}
\label{tab:multiTask}
\end{table}
We are also interested in exploring the relationship among multiple pretext tasks. We can see from Table \ref{tab:multiTask} that certain pairs of tasks performs better than the others, such as \textbf{{\em  speed}} with \textbf{{\em shuffle}}, \textbf{{\em  reverse}}  with \textbf{{\em  rotation jittering}} .
Here, in order to obtain a more holistic view of what makes good combinations among the pretext tasks, we first pretrain each pretext task respectively on Kinetics-400 dataset, then finetune with other pretext tasks on UCF-101. 
Specifically, in Figure \ref{fig:pretext-task}, we pretrain on the row's pretext tasks and finetune on the column's pretext tasks. The numbers in Figure \ref{fig:pretext-task} indicate the finetuning accuracy on UCF-101. 
We observe \textbf{{\em rotation}} and \textbf{{\em shuffle}} seem to perform better with the \textbf{{\em reverse}}  and \textbf{{\em speed}} , respectively. 
Also combining  \textbf{{\em speed}} task with \textbf{{\em shuffle}} task yields the highest accuracy (0.968) in Figure \ref{fig:pretext-task}. 
This is further validated in Table \ref{tab:multiTask} that TaCo with \textbf{{\em  speed}}  and \textbf{{\em  shuffle}}  tasks perform the best on both datasets, \ie, 78.77\% accuracy for UCF-101 and 40.62\% accuracy on HMDB-51. Our conjecture is that, \textbf{{\em  shuffle}}  and \textbf{{\em  speed}}  tasks mainly focus on subtle motion changes while \textbf{{\em  rotation jittering}}  and \textbf{{\em  reverse}}  tasks focus more on the global transformation.
We provide Figure \ref{fig:teaser} to illustrate this in a more intuitive way. 
Given different pretext tasks, we use RAFT \cite{teed2020raft} to predict the optical flow of two different cases (as shown in the first and the second row). Then we substract these two flow fields and visualize their motion difference (the third row). 
We observe that, for \textbf{{\em  rotation jittering}}  and \textbf{{\em  reverse}} , the motion difference between each case is large, indicating that these two tasks favor global information in TaCo. While for \textbf{{\em  speed}}  and \textbf{{\em  shuffle}} , their motion difference appears to be relatively small, which implies that these tasks should attend to local cues instead.

Besides the dual-task setting, we also perform experiments on triple or even quadruple tasks, which does not yield better results than the results achieved under the dual-task setting. This is largely due to that these tasks are not highly harmonized, as we have discussed above.

\subsection{Experimental Results}
\paragraph{Generalization.} To show the generalization ability of TaCo, we further provide results of TaCo under different settings, including different contrastive learning frameworks and backbone architectures. As shown in Table \ref{tab:differentsetting}, our results suggest a solid and consistent improvement among all settings.
\begin{table}[h]
\centering
\footnotesize
\begin{tabular}{c|c|c|c|c}
\shline
\multicolumn{1}{l|}{Frame} & Backbone                   & CSL      & UCF           & HMDB          \\ \shline
\multirow{6}{*}{8}          & \multirow{2}{*}{R18}       & InstDisc & 78.8          & 40.6          \\ \cline{3-5} 
                            &                            & MoCo     & 81.0          & 42.2          \\ \cline{2-5} 
                            & \multirow{2}{*}{R(2+1)D18} & InstDisc & 79.1          & 38.2          \\ \cline{3-5} 
                            &                            & MoCo     & 80.6          & 40.0          \\ \cline{2-5} 
                            & \multirow{2}{*}{R50}       & InstDisc & 81.6          & 44.2          \\ \cline{3-5} 
                            &                            & MoCo     & 83.9          & 45.3          \\ \hline
\multirow{6}{*}{16}         & \multirow{2}{*}{R18}       & InstDisc & 79.1          & 44.6          \\ \cline{3-5} 
                            &                            & MoCo     & 81.4          & 45.4          \\ \cline{2-5} 
                            & \multirow{2}{*}{R(2+1)D18} & InstDisc & 80.7          & 43.3          \\ \cline{3-5} 
                            &                            & MoCo     & 81.8          & 46.0          \\ \cline{2-5} 
                            & \multirow{2}{*}{R50}       & InstDisc & 83.9          & 48.7          \\ \cline{3-5} 
                            &                            & MoCo     & \textbf{85.1} & \textbf{51.6} \\ \hline
\end{tabular}
\caption{Top-1 accuracy (\%) of the classification results on UCF-101 and HMDB-51 of TaCo under different settings.}
\label{tab:differentsetting}
\end{table}
\paragraph{Comparison with other self-supervised methods.}
As shown in Table~\ref{tab:benchmark}, features learned by TaCo successfully transfer to other downstream video tasks on UCF-101 and HMDB-51, and outperforms other unsupervised video representation learning methods by a large margin.

\begin{table}[]
\footnotesize
\centering
\begin{tabular}{l|c|c|c}
\shline
Method       & Backbone   & UCF                            & HMDB                           \\ \shline
MotionPred\cite{wang2019self} & C3D        & 61.2                           & 33.4                           \\ \hline
3DRotNet\cite{jing2018self}   & R18        & 62.9                           & 33.7                           \\ \hline
ST-Puzzle\cite{kim2019self}  & R18        & 65.8                           & 33.7                           \\ \hline
ClipOrder\cite{xu2019self}  & R(2+1)D-18 & 66.7                           & 43.7                           \\ \hline
DPC\cite{han2019video}        & R34        & 72.4                           & 30.9                           \\ \hline
AoT\cite{wei2018learning}       & T-CAM      & 75.7                           & 35.7                           \\ \hline
SpeedNet\cite{benaim2020speednet}   & I3D        & 66.7                           & 43.7                           \\ \hline
PacePred\cite{wang2020self}   & R(2+1)D-18 & 77.1                           & 36.6                           \\ \hline
MemDPC\cite{han2020memory}   & R-2D3D     & 78.1                           & 41.2                           \\ \hline
CBT\cite{sun2019learning} & S3D & 79.5 &44.6 \\ \hline
VTHCL\cite{yang2020video}      & R50        & 82.1                           & 49.2                           \\ \shline
TaCo(MoCo)                         & R50        & \textbf{85.1} & \textbf{51.6} \\ \hline
\end{tabular}
\caption{Top-1 accuracy (\%) on UCF-101 and HMDB-51.}
\label{tab:benchmark}
\end{table}
\section{Ablation Study}
\paragraph{Pre-training using task losses only.}

In Table \ref{tab:singleTask}, by comparing with 3D InstDisc, we show that TaCo can significantly improve the unsupervised video representation learning via introducing additional task losses. 
Here, we would like to show the effectiveness of contrastive learning by disabling the contrastive loss, \ie, only optimizing the task losses for pre-training. It is worth noting that TaCo is modified to better suit the contrastive learning framework, thus the settings and implementations are not the same as the original works which we compared in Table \ref{tab:benchmark}. 
From Table \ref{tab:taskonly}, we observe that the optimizing task losses only performs worse than TaCo, which shows the importance of the contrastive loss. 
It is also worth mentioning that our previous analysis regarding the relationship among different tasks (Section \ref{sec:expresult}) also holds true in the current setting, \ie, ``\textbf{{\em  rotation jittering}}  + \textbf{{\em  reverse}} '' and ``\textbf{{\em  speed}}  + \textbf{{\em  shuffle}} '' are the best among all combinations. 
\begin{table}[h]
\footnotesize
\centering
\begin{tabular}{c|c|c|c}
\shline
 & \multicolumn{1}{c|}{Method}  & UCF & HMDB \\ \shline
\multirow{4}{*}{\begin{tabular}[c]{@{}l@{}}Single-\\ Task\end{tabular}} & \multicolumn{1}{c|}{Rotation} & 61.92 & 32.60 \\ \cline{2-4} 
 & \multicolumn{1}{c|}{Reverse} & 60.86   & 33.31    \\ \cline{2-4} 
 & \multicolumn{1}{c|}{Shuffle}  & 62.36   &  30.61    \\ \cline{2-4} 
 & \multicolumn{1}{c|}{Speed}   & 62.01   & 31.01   \\\shline
\multirow{6}{*}{\begin{tabular}[c]{@{}l@{}}Multi-\\ Task\end{tabular}}  & Rotation+Reverse              & 63.01 & 35.19 \\ \cline{2-4} 
 & Rotation+Shuffle              & 62.48   & 32.84    \\ \cline{2-4} 
 & Rotation+Speed               & 62.69   & 34.13    \\ \cline{2-4} 
 & Shuffle+Reverse               & 63.88   & 31.17    \\ \cline{2-4} 
 & Speed+Shuffle                 & 64.12   & 35.01    \\ \cline{2-4} 
 & Reverse+Speed                & 62.97   & 34.66    \\ \hline
\end{tabular}
\caption{Top-1 accuracy (\%) on UCF-101 and HMDB-51 with only applying temporal transformation tasks without contrastive learning. These temporal tasks are analysed in the same setting as above.}
\label{tab:taskonly}
\end{table}

\paragraph{The importance of $\lambda$.}

There are multiple loss terms in TaCo (Equation~\ref{eqn:overall-loss}) which need to be balanced for a better coordination.
As aforementioned in Section~\ref{sec:method}, we need a relatively larger value for the balance parameter $\lambda$ so as to prevent the task-related loss from being overwhelmed.
In our previous experiments, we set $\lambda = 10$ since we notice that the initial numerical value of $L_{contrast}$ is about $10\times$ as large as that of $L_{task}$. 
Here we also provide an ablation analysis to study the importance of $\lambda$. 
By varying the value of $\lambda$ from 0 to 20, the results of TaCo using \textbf{{\em  rotation jittering}}   as the temporal pretext task are reported in Table \ref{tab:lambda}. We observe that $\lambda = 10$ achieves the best performance. Meanwhile, we note that TaCo is not sensitive to this parameter when $\lambda$ is drawn from $10\sim15$.

\begin{table}[h]
\footnotesize
\centering
\begin{tabular}{l|l|l|l|l|l}
\shline
     & $\lambda$=0 & $\lambda$=5 & $\lambda$=10 & $\lambda$=15 & $\lambda$=20 \\ \shline
TaCo(Rot) &  70.33  &  72.90  &  73.81   &  73.74   & 73.06   \\ \hline
\end{tabular}
\caption{Top-1 accuracy (\%) on UCF-101. Temporal \textbf{{\em  rotation jittering}}   task is applied. $\lambda$ indicates the hyper-parameter used to balance the contrastive loss and task loss.}
\label{tab:lambda}
\end{table}

\begin{table}[h]
\footnotesize
\centering
\begin{tabular}{c|c|c|c}
\shline
\multicolumn{2}{c|}{\multirow{2}{*}{Method}} &
  \multicolumn{2}{c}{Last Layer} \\ \cline{3-4} 
\multicolumn{2}{c|}{}              & UCF   & HMDB   \\ \shline
\multicolumn{2}{c|}{CSL: 3D InstDisc} & 53.52          & 23.82         \\ \shline
\multirow{4}{*}{\begin{tabular}[c]{@{}c@{}} Temporal\\ Augment \end{tabular}} &
  Rotation &
  54.60 &
  22.72\\ \cline{2-4} 
              & Reverse             & 54.13          & 24.91        \\ \cline{2-4} 
              & Shuffle              & 51.64          & 22.38           \\ \cline{2-4} 
              & Speed               & 53.87          & 23.92          \\ \shline
\multirow{4}{*}{\begin{tabular}[c]{@{}c@{}}TaCo\\ (Ours)\end{tabular}} &
  Rotation &
  57.70 &
  26.49 \\ \cline{2-4} 
              & Reverse             & \textbf{59.63}          & \textbf{26.70}           \\ \cline{2-4} 
              & Shuffle              & 54.03          & 25.21          \\ \cline{2-4} 
              & Speed               & 58.01          & 26.19          \\ \hline
\end{tabular}
\caption{Top-1 accuracy (\%) on UCF-101 and HMDB-51 of linear evaluation protocol.}
\label{tab:lastlayer}
\end{table}
\paragraph{Last layer finetuning.}
We also perform linear evaluation on UCF-101 and HMDB-51. Experimental results in Table \ref{tab:lastlayer} suggest that TaCo also demonstrates a solid improvement under the linear evalutation setting, compared to vanilla CSL or temporal augmentations alone. 

\section{Conclusion}
\label{sec:conclusion}
In this paper we present Temporal-aware Contrastive self-supervised learning (TaCo), a general framework to enhance video CSL. TaCo is motivated by the counter-intuitive observation that directly applying temporal augmentations does not help. TaCo enables effective integration of temporal information by selecting temporal transformations not only as strong augmentation but also to constitute extra self-supervision under CSL paradigm. As a generic and flexible framework, TaCo can well accommodate various temporal transformations, backbones and CSL approaches. It also delivers competitive results on action classification benchmarks. In the future, we plan to apply TaCo to other downstream tasks and to investigate other temporal self-supervision signals. 

\section*{Acknowledgement}
We would like to thank Christoph Feichtenhofer for the discussion and valuable feedbacks.\\

{\small
\bibliographystyle{ieee_fullname}
\bibliography{egbib}

\begin{thebibliography}{10}\itemsep=-1pt

\bibitem{benaim2020speednet}
Sagie Benaim, Ariel Ephrat, Oran Lang, Inbar Mosseri, William~T Freeman,
  Michael Rubinstein, Michal Irani, and Tali Dekel.
\newblock Speednet: Learning the speediness in videos.
\newblock In {\em Proceedings of the IEEE/CVF Conference on Computer Vision and
  Pattern Recognition}, pages 9922--9931, 2020.

\bibitem{chen2020simple}
Ting Chen, Simon Kornblith, Mohammad Norouzi, and Geoffrey Hinton.
\newblock A simple framework for contrastive learning of visual
  representations.
\newblock {\em arXiv preprint arXiv:2002.05709}, 2020.

\bibitem{doersch2015unsupervised}
Carl Doersch, Abhinav Gupta, and Alexei~A. Efros.
\newblock Unsupervised visual representation learning by context prediction.
\newblock In {\em ICCV}, 2015.

\bibitem{doersch2017multi}
Carl Doersch and Andrew Zisserman.
\newblock Multi-task self-supervised visual learning.
\newblock In {\em Proceedings of the IEEE International Conference on Computer
  Vision}, pages 2051--2060, 2017.

\bibitem{fan2020pyslowfast}
Haoqi Fan, Yanghao Li, Bo Xiong, Wan-Yen Lo, and Christoph Feichtenhofer.
\newblock Pyslowfast.
\newblock \url{https://github.com/facebookresearch/slowfast}, 2020.

\bibitem{feichtenhofer2019slowfast}
Christoph Feichtenhofer, Haoqi Fan, Jitendra Malik, and Kaiming He.
\newblock Slowfast networks for video recognition.
\newblock In {\em Proceedings of the IEEE international conference on computer
  vision}, pages 6202--6211, 2019.

\bibitem{fernando2017self}
Basura Fernando, Hakan Bilen, Efstratios Gavves, and Stephen Gould.
\newblock Self-supervised video representation learning with odd-one-out
  networks.
\newblock In {\em Proceedings of the IEEE conference on computer vision and
  pattern recognition}, pages 3636--3645, 2017.

\bibitem{gidaris2018unsupervised}
Spyros Gidaris, Praveer Singh, and Nikos Komodakis.
\newblock Unsupervised representation learning by predicting image rotations.
\newblock {\em ICLR}, 2018.

\bibitem{gutmann2010noise}
Michael Gutmann and Aapo Hyv{\"a}rinen.
\newblock Noise-contrastive estimation: A new estimation principle for
  unnormalized statistical models.
\newblock In {\em Proceedings of the Thirteenth International Conference on
  Artificial Intelligence and Statistics}, pages 297--304, 2010.

\bibitem{han2019video}
Tengda Han, Weidi Xie, and Andrew Zisserman.
\newblock Video representation learning by dense predictive coding.
\newblock In {\em ICCV Workshops}, 2019.

\bibitem{han2020memory}
Tengda Han, Weidi Xie, and Andrew Zisserman.
\newblock Memory-augmented dense predictive coding for video representation
  learning.
\newblock {\em arXiv preprint arXiv:2008.01065}, 2020.

\bibitem{he2020momentum}
Kaiming He, Haoqi Fan, Yuxin Wu, Saining Xie, and Ross Girshick.
\newblock Momentum contrast for unsupervised visual representation learning.
\newblock In {\em Proceedings of the IEEE/CVF Conference on Computer Vision and
  Pattern Recognition}, pages 9729--9738, 2020.

\bibitem{jing2018self}
Longlong Jing, Xiaodong Yang, Jingen Liu, and Yingli Tian.
\newblock Self-supervised spatiotemporal feature learning via video rotation
  prediction.
\newblock {\em arXiv preprint arXiv:1811.11387}, 2018.

\bibitem{kay2017kinetics}
Will Kay, Joao Carreira, Karen Simonyan, Brian Zhang, Chloe Hillier, Sudheendra
  Vijayanarasimhan, Fabio Viola, Tim Green, Trevor Back, Paul Natsev, et~al.
\newblock The kinetics human action video dataset.
\newblock {\em arXiv preprint arXiv:1705.06950}, 2017.

\bibitem{kim2019self}
Dahun Kim, Donghyeon Cho, and In~So Kweon.
\newblock Self-supervised video representation learning with space-time cubic
  puzzles.
\newblock In {\em AAAI}, 2019.

\bibitem{kuehne2011hmdb}
Hildegard Kuehne, Hueihan Jhuang, Est{\'\i}baliz Garrote, Tomaso Poggio, and
  Thomas Serre.
\newblock Hmdb: a large video database for human motion recognition.
\newblock In {\em ICCV}, 2011.

\bibitem{misra2020self}
Ishan Misra and Laurens van~der Maaten.
\newblock Self-supervised learning of pretext-invariant representations.
\newblock In {\em Proceedings of the IEEE/CVF Conference on Computer Vision and
  Pattern Recognition}, pages 6707--6717, 2020.

\bibitem{noroozi2016unsupervised}
Mehdi Noroozi and Paolo Favaro.
\newblock Unsupervised learning of visual representations by solving jigsaw
  puzzles.
\newblock In {\em ECCV}, 2016.

\bibitem{noroozi2017representation}
Mehdi Noroozi, Hamed Pirsiavash, and Paolo Favaro.
\newblock Representation learning by learning to count.
\newblock In {\em ICCV}, 2017.

\bibitem{cpc}
Aaron van~den Oord, Yazhe Li, and Oriol Vinyals.
\newblock Representation learning with contrastive predictive coding.
\newblock {\em arXiv preprint arXiv:1807.03748}, 2018.

\bibitem{purushwalkam2020aligning}
Senthil Purushwalkam, Tian Ye, Saurabh Gupta, and Abhinav Gupta.
\newblock Aligning videos in space and time.
\newblock {\em arXiv preprint arXiv:2007.04515}, 2020.

\bibitem{qian2020spatiotemporal}
Rui Qian, Tianjian Meng, Boqing Gong, Ming-Hsuan Yang, Huisheng Wang, Serge
  Belongie, and Yin Cui.
\newblock Spatiotemporal contrastive video representation learning.
\newblock {\em arXiv preprint arXiv:2008.03800}, 2020.

\bibitem{soomro2012ucf101}
Khurram Soomro, Amir~Roshan Zamir, and Mubarak Shah.
\newblock Ucf101: A dataset of 101 human actions classes from videos in the
  wild.
\newblock {\em arXiv preprint arXiv:1212.0402}, 2012.

\bibitem{sun2019learning}
Chen Sun, Fabien Baradel, Kevin Murphy, and Cordelia Schmid.
\newblock Learning video representations using contrastive bidirectional
  transformer.
\newblock {\em arXiv preprint arXiv:1906.05743}, 2019.

\bibitem{taleb20203d}
Aiham Taleb, Winfried Loetzsch, Noel Danz, Julius Severin, Thomas Gaertner,
  Benjamin Bergner, and Christoph Lippert.
\newblock 3d self-supervised methods for medical imaging.
\newblock {\em NeurIPS}, 2020.

\bibitem{teed2020raft}
Zachary Teed and Jia Deng.
\newblock Raft: Recurrent all-pairs field transforms for optical flow.
\newblock {\em arXiv preprint arXiv:2003.12039}, 2020.

\bibitem{tian2019contrastive}
Yonglong Tian, Dilip Krishnan, and Phillip Isola.
\newblock Contrastive multiview coding.
\newblock {\em arXiv preprint arXiv:1906.05849}, 2019.

\bibitem{wang2019self}
Jiangliu Wang, Jianbo Jiao, Linchao Bao, Shengfeng He, Yunhui Liu, and Wei Liu.
\newblock Self-supervised spatio-temporal representation learning for videos by
  predicting motion and appearance statistics.
\newblock In {\em CVPR}, 2019.

\bibitem{wang2020self}
Jiangliu Wang, Jianbo Jiao, and Yun-Hui Liu.
\newblock Self-supervised video representation learning by pace prediction.
\newblock {\em arXiv preprint arXiv:2008.05861}, 2020.

\bibitem{wang2018non}
Xiaolong Wang, Ross Girshick, Abhinav Gupta, and Kaiming He.
\newblock Non-local neural networks.
\newblock In {\em Proceedings of the IEEE conference on computer vision and
  pattern recognition}, pages 7794--7803, 2018.

\bibitem{CVPR2019_CycleTime}
Xiaolong Wang, Allan Jabri, and Alexei~A. Efros.
\newblock Learning correspondence from the cycle-consistency of time.
\newblock In {\em CVPR}, 2019.

\bibitem{wei2018learning}
Donglai Wei, Joseph~J Lim, Andrew Zisserman, and William~T Freeman.
\newblock Learning and using the arrow of time.
\newblock In {\em CVPR}, 2018.

\bibitem{wu2018unsupervised}
Zhirong Wu, Yuanjun Xiong, Stella~X Yu, and Dahua Lin.
\newblock Unsupervised feature learning via non-parametric instance
  discrimination.
\newblock In {\em Proceedings of the IEEE Conference on Computer Vision and
  Pattern Recognition}, pages 3733--3742, 2018.

\bibitem{xu2019self}
Dejing Xu, Jun Xiao, Zhou Zhao, Jian Shao, Di Xie, and Yueting Zhuang.
\newblock Self-supervised spatiotemporal learning via video clip order
  prediction.
\newblock In {\em CVPR}, 2019.

\bibitem{yang2020video}
Ceyuan Yang, Yinghao Xu, Bo Dai, and Bolei Zhou.
\newblock Video representation learning with visual tempo consistency.
\newblock {\em arXiv preprint arXiv:2006.15489}, 2020.

\bibitem{zhang2016colorful}
Richard Zhang, Phillip Isola, and Alexei~A Efros.
\newblock Colorful image colorization.
\newblock In {\em ECCV}, 2016.

\end{thebibliography}
}

\clearpage
\appendix

\section{Last-layer Finetuning}
\begin{figure}[h]
    \centering
    \includegraphics[width=0.75\linewidth]{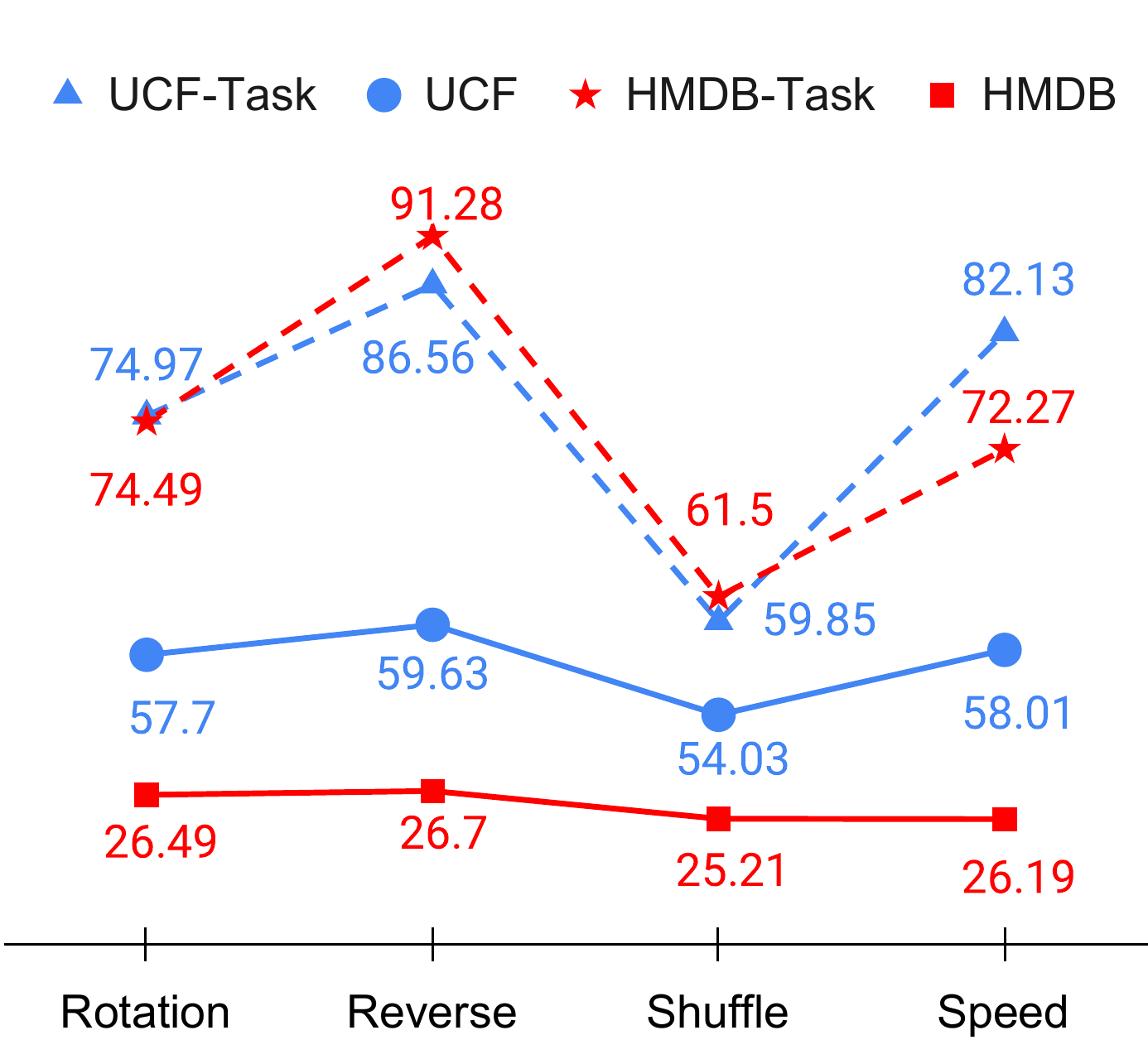}
    \caption{TaCo with different pretext tasks (solid lines) vs. CSL models finetuned on different pretext tasks (dash lines).
    }
    \label{fig:singletrend}
\end{figure}

To further analyze the correlation between the pretext tasks and contrastive learning, we use contrastive learning as pretraining and then finetune \emph{the last layer} on different pretext tasks, similar as the analysis for all-layer finetuning in Figure~\ref{fig:trend}.  Figure \ref{fig:singletrend} shows the finetuning top-1 accuracy of different video pretext tasks on UCF-101 and HMDB-51, pretrained with 3D InstDisc under ResNet-18, 8-frame settings (in dash lines). As for comparison, the results of TaCo with different pretext tasks are also demonstrated (in solid lines).

As is shown in Figure \ref{fig:singletrend}, our finding in Section~\ref{sec:expresult} still holds for linear finetuning: a higher finetuning performance indicates a better pretext task to be employed. 

Though under different finetuning settings, the best-performing task with TaCo can be different,  our simple strategy and analysis can help choose the best task to deploy with TaCo in a more efficient way.
\section{Using More Pretext Tasks}
Besides  the  dual-task  setting,  we  also  perform  experiments  on  triple  or  even  quadruple  tasks (see Table \ref{tab:multiTask}),  which  does  not yield better results than the results achieved under the dual-task setting. This is largely because these tasks are not harmonized, as we have discussed in Section~\ref{sec:expresult}.

\begin{table}[]
\centering
\footnotesize
\begin{tabular}{c|c|c|c|c}
\shline
\multicolumn{3}{c|}{Method}        & UCF            &HMDB\\ \shline
\multicolumn{3}{c|}{CSL: 3D InstDisc} & 70.33          & 33.23          \\ \shline
\multirow{4}{*}{\begin{tabular}[c]{@{}c@{}}Single \\Task\end{tabular}} &
  \multirow{4}{*}{\begin{tabular}[c]{@{}c@{}}TaCo\end{tabular}} &
  Rotation &
  73.81 &
  36.88 \\ \cline{3-5} 
       &      & Reverse             & 74.57          & \textbf{37.23} \\ \cline{3-5} 
       &      & Shuffle              & \textbf{76.02} & 35.59          \\ \cline{3-5} 
       &      & Speed               & 74.38          & 36.13          \\ \shline
\multirow{6}{*}{\begin{tabular}[c]{@{}c@{}}Dual\\ Task\end{tabular}} &
  \multirow{6}{*}{\begin{tabular}[c]{@{}c@{}}TaCo\end{tabular}} &
  Rotation+Reverse &
  75.24 &
  39.02 \\ \cline{3-5} 
       &      & Rotation+Shuffle     & 74.92          & 37.44          \\ \cline{3-5} 
       &      & Rotation+Speed      & 75.03          & 38.15          \\ \cline{3-5} 
       &      & Shuffle+Reverse      & 75.69          & 36.54          \\ \cline{3-5} 
       &      & Speed+Shuffle        & \textbf{78.77} & \textbf{40.62} \\ \cline{3-5} 
       &      & Reverse+Speed       & 75.18          & 38.96          \\ \shline
\multirow{4}{*}{\begin{tabular}[c]{@{}c@{}}Triple \\Task\end{tabular}} &
  \multirow{4}{*}{\begin{tabular}[c]{@{}c@{}}TaCo\end{tabular}} &
  Rotation+Speed+Shuffle &
  76.54 &
  38.71 \\ \cline{3-5}
  &      & Reverse+Speed+Shuffle       & 77.22  & 39.10  \\ \cline{3-5}
  &      & Shuffle+Rotation+Reverse       & 74.70  & 37.49  \\ \cline{3-5}
  &      & Speed+Rotation+Reverse       & 75.28  & 37.80 \\ \shline
\multirow{2}{*}{\begin{tabular}[c]{@{}c@{}} Quadruple \\ Task\end{tabular}} &
  \multirow{2}{*}{\begin{tabular}[c]{@{}c@{}}TaCo \end{tabular}} &
  \multirow{2}{*}{\begin{tabular}[c]{@{}c@{}} Rotation+Reverse+\\ Speed+Shuffle\end{tabular}} &
  \multirow{2}{*}{\begin{tabular}[c]{@{}c@{}} 75.88 \end{tabular}} &
  \multirow{2}{*}{\begin{tabular}[c]{@{}c@{}} 37.27 \end{tabular}} \\ 
  & & & & \\ \hline
\end{tabular}
\caption{{Different combination of video pretext tasks:} top-1 accuracy (\%) of classification results on UCF-101 and HMDB-51 with fully finetuned backbone of TaCo.}

\label{tab:multiTask}
\end{table}

\section{Other Pretext Tasks}
In Section~\ref{sec:expresult} we mentioned that not all pretext tasks are equally suited for video CSL. Under the 8-frame setting, our empirical results show a clear performance drop of TaCo by using ClipOrder \cite{xu2019self}.  Our hypothesis is that 8 frames are too short to provide useful temporal information for solving the re-ordering task, 
while re-ordering longer clips may result in substantially heavier computation overhead, which can be impractical in video CSL.

\begin{table}[h]
\footnotesize
\centering
\begin{tabular}{c|c|c|c}
\shline
\multicolumn{2}{c|}{\multirow{2}{*}{Method}} &
  \multicolumn{2}{c}{Fully Finetune} \\ \cline{3-4} 
\multicolumn{2}{c|}{}              & UCF   & HMDB\\ \shline
\multicolumn{2}{c|}{baseline (3D InstDisc)} &  70.33          & 33.23 \\ \shline
\multirow{4}{*}{\begin{tabular}[c]{@{}c@{}}TaCo\end{tabular}} &
  Rotation &
  73.81 &
  36.89\\ \cline{2-4} 
              & Reverse             & 74.57 & \textbf{37.23} \\ \cline{2-4} 
              & Shuffle              & \textbf{76.02} & 35.59 \\ \cline{2-4} 
              & Speed                & 74.38          & 36.13 \\ \shline
\multirow{1}{*}{\begin{tabular}[c]{@{}c@{}}TaCo\end{tabular}} &
  ClipOrder & 66.83
   & 29.05
  \\ \hline
\end{tabular}
\caption{Comparison among different video pretext tasks: top-1 accuracy (\%) on UCF-101 and HMDB-51 classification of TaCo using different pretext tasks with fully finetuning schedule.}
\end{table}
\end{document}